\documentclass[runningheads]{llncs}

\usepackage{eccv}

\usepackage{eccvabbrv}

\usepackage{graphicx}
\usepackage{booktabs}
\usepackage{caption}
\usepackage[accsupp]{axessibility}  

\usepackage{hyperref}

\usepackage{orcidlink}
\usepackage{wrapfig}

\newcommand{\Imat}{{\boldsymbol I}}

\newcommand{\Kmat}[0]{{{\boldsymbol K}}}

\newcommand{\Mmat}[0]{{{\boldsymbol M}}}

\newcommand{\Qmat}[0]{{{\boldsymbol Q}}}

\newcommand{\Vmat}[0]{{{\boldsymbol V}}}
\newcommand{\Wmat}[0]{{{\boldsymbol W}}}
\newcommand{\Xmat}{{\boldsymbol X}}
\newcommand{\Ymat}[0]{{{\boldsymbol Y}}}

\newcommand{\Thetamat}{\boldsymbol{\Theta}}

\newcommand{\Sigmamat}{\boldsymbol{\Sigma}}

\newcommand{\mysubscript}[1]{\raisebox{-0.34ex}{\scriptsize#1}}

\begin{document}

\title{Hierarchical Separable Video Transformer for Snapshot Compressive Imaging} 


\author{Ping Wang\inst{1,2}\orcidlink{0009-0001-2746-5102} \and
Yulun Zhang\inst{3}\orcidlink{0000-0002-2288-5079} \and
Lishun Wang\inst{2}\orcidlink{0000-0003-3245-9265}\and
Xin Yuan\inst{2}\thanks{Corresponding author.}\orcidlink{0000-0002-8311-7524}}

\authorrunning{P.~Wang et al.}

\institute{Zhejiang University, Hangzhou, China \and Westlake University, Hangzhou, China
\and
Shanghai Jiao Tong University, Shanghai, China\\
\email{\{wangping,wanglishun,xyuan\}@westlake.edu.cn} ~~~\email{yulun100@gmail.com}
}

\maketitle

\begin{abstract}
Transformers have achieved the state-of-the-art performance on solving the inverse problem of Snapshot Compressive Imaging (SCI) for video, whose ill-posedness is rooted in the mixed degradation of spatial masking and temporal aliasing.
However, previous Transformers lack an insight into the degradation and thus have limited performance and efficiency.
In this work, we tailor an efficient reconstruction architecture without temporal aggregation in early layers and Hierarchical Separable Video Transformer (HiSViT) as building block.
HiSViT is built by multiple groups of Cross-Scale Separable Multi-head Self-Attention (CSS-MSA) and Gated Self-Modulated Feed-Forward Network (GSM-FFN) with dense connections, each of which is conducted within a separate channel portions at a different scale, for multi-scale interactions and long-range modeling.
By separating spatial operations from temporal ones, CSS-MSA introduces an inductive bias of paying more attention within frames instead of between frames while saving computational overheads.
GSM-FFN further enhances the locality via gated mechanism and factorized spatial-temporal convolutions.
Extensive experiments demonstrate that our method outperforms previous methods by $\!>\!0.5$ dB with comparable or fewer parameters and complexity.
The source codes and pretrained models are released at \url{https://github.com/pwangcs/HiSViT}.
\end{abstract}

\section{Introduction}
\label{sec:intro}
High-speed cameras are crucial vision devices for scientific research, industrial manufacturing, and environmental monitoring.
Unlike typical expensive high-speed cameras, Snapshot Compressive Imaging (SCI)~\cite{YuanSCI,reddy2011p2c2,Hitomi11_ICCV_videoCS,llull2013coded,gao2014single,koller2015high,martel2020neural,wang2023full} multiplexes a sequence of video frames, each of which is optically modulated with temporally-varying masks, into a single-shot observation of a low-cost monochromatic camera for high speed and low storage.
Optical modulation and multiplexing lead to two corresponding degradations: spatial masking and temporal aliasing.
Similar to compressive sensing problems~\cite{duarte2008single,sun2016deep,wang2023saunet}, the inverse problem of video SCI is to reconstruct multiple high-fidelity frames from the observed image.
As demonstrated in~\cref{fig:pipeline} (a), multiple frames are first initialized from the observed image and known masks and then they are input to an optimization algorithm or a deep model for effective restoration.
In this context, video SCI reconstruction can be viewed as a challenging video restoration task, like denoising, deblurring, \etc.
Actually, they are vastly different in data distribution.
As depicted in~\cref{fig:pipeline} (b), input frames of video SCI reconstruction lose temporal correlations (\ie, motion dynamics) completely due to the mixed degradation of spatial masking and temporal aliasing, differing from that input frames of a plain video restoration task are highly-related with clear frames even degraded.
For video SCI reconstruction, informative clues concentrate on spatial dimensions as opposed to temporal dimension, referred to as {\em information skewness}.

\begin{figure}[t]
\centering
\includegraphics[width=1\linewidth]{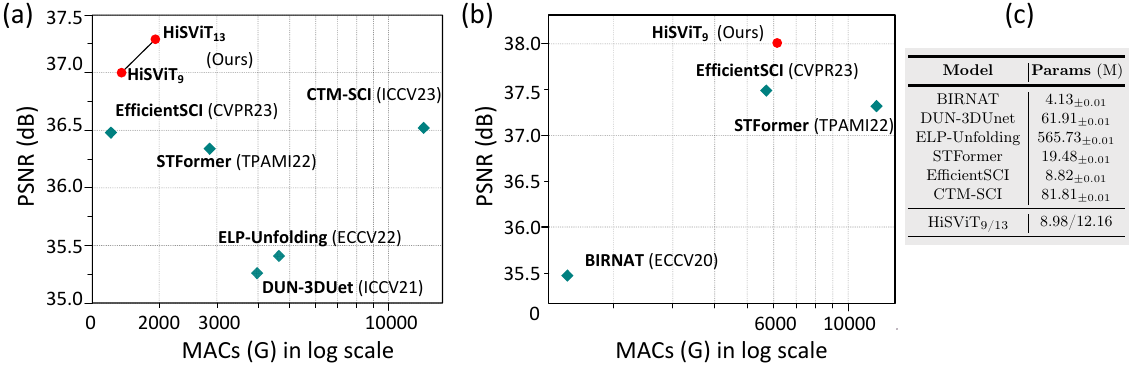}
\caption{Our HiSViT achieves SOTA performance on (a) grayscale and (b) color video SCI reconstruction with comparable or fewer MACs and (c) parameters.}
\label{fig:performance}
\end{figure}

Video SCI reconstruction has been extensively studied under straight~\cite{Cheng2021_CVPR_ReverSCI,Wang2021_CVPR_MetaSCI,wang2022spatial,wang2023efficientsci}, U-shaped~\cite{qiao2020deep,wang2023deep}, recurrent~\cite{Cheng20ECCV_Birnat,cheng2022recurrent}, unrolling~\cite{Ma19ICCV,wu2021dense,yang2022ensemble,meng2023deep,Zheng_2023_ICCV}, and plug-and-play~\cite{Yuan2020_CVPR_PnP,yuan2021plug} architectures.
From early convolutional models~\cite{Ma19ICCV,qiao2020deep,Yuan2020_CVPR_PnP,yuan2021plug,Cheng2021_CVPR_ReverSCI,wu2021dense,yang2022ensemble,meng2023deep,wang2023deep} to latest Transformer models~\cite{wang2022spatial,wang2023efficientsci,Zheng_2023_ICCV}, the performance gains are due in large part to advanced vision engines and they lack an insight into the information skewness.
Due to the limited perception field and static kernel of convolution, CNN-based models have inherent shortcomings in capturing long-range dependencies and learning generalizable priors.
Recently, Transformer-based models~\cite{wang2022spatial,wang2023efficientsci,Zheng_2023_ICCV} have achieved the state-of-the-art (SOTA) performance.
As a core component of Transformer, Multi-head Self-Attention (MSA) mechanism is highly effective in capturing long-range dependencies by aggregating all tokens weighted by the similarity between them.
However, vanilla Global MSA (G-MSA)~\cite{dosovitskiy2020image} suffers from the quadratic computational complexity with respect to token numbers, thus being impractical for high-dimensional data, like video.
To relieve computational loads, STFormer~\cite{wang2022spatial}, a variant of Factorized MSA (F-MSA)~\cite{bertasius2021space,arnab2021vivit}, applies 2D Windowed MSA (W-MSA)~\cite{liu2021swin} on spatial dimensions and 1D G-MSA on temporal dimension in a 
separate and parallel manner to surpass CNN-based models.
By replacing spatial W-MSA of STFormer with 2D convolutions, EfficientSCI~\cite{wang2023efficientsci} further improves the performance with less computational loads.
STFormer and EfficientSCI don't conduct MSA on video space directly, thus they are not real video Transformers.
CTM-SCI~\cite{Zheng_2023_ICCV} first applies 3D W-MSA~\cite{liu2022video} on video space in an unrolling architecture to get the latest result (36.52 dB) at the cost of extremely high computational loads (12.79 TMACs).
By rethinking the information skewness and Transformers' gain, we point out the keys of video SCI reconstruction:
{$i)$ spatial aggregation plays a more important role than temporal one}, and {$ii)$ long-range spatial-temporal modeling is desired but usually at the expense of high computational complexity}.

In this work, we make several modifications on reconstruction architecture and Transformer block to fulfill the above requirements.
Towards architecture, previous models generally apply stacked 3D convolutions to transform degraded frames into shallow features.
In absence of motion dynamics, temporal interactions in early layers could exaggerate artifacts owing to error accumulation during propagation~\cite{chan2022investigating}, leading to poor representations (see~\cref{fig:feature}). To this end, we appeal for using 2D operator as the frame-wise shallow feature extractor. 
Towards building block, we propose an efficient Hierarchical Separable Video Transformer (HiSViT) to tackle the mixed degradation of video SCI, powered by Cross-Scale Separable Multi-head Self-Attention (CSS-MSA) and Gated Self-Modulated Feed-Forward Network (GSM-FFN). 
CSS-MSA, a spatial-then-temporal attention, separates spatial operations from temporal ones within a single attention layer.
Such separation design leads to: $i)$ computational efficiency; $ii)$ an inductive bias of paying more attention within frames instead of between frames.
The former is similar to previous F-MSA~\cite{bertasius2021space,arnab2021vivit} by breaking the direct interactions between non-aligned tokens, located at both different frames and different spatial locations.
The later is customized to harmonize with the information skewness of video SCI.
Besides, spatial receptive field is designed to be windowed yet increasing along heads for efficient multi-scale representation learning and temporal receptive field is global considering the limited frames to be processed as demonstrated in~\cref{fig:hisvit} (d).
GSM-FFN could strengthen the locality by introducing gated self-modulation and factorized Spatial-Temporal Convolution (STConv) to regular FFN.
Each HiSViT block is built by multiple groups of CSS-MSA and GSM-FFN with dense connections, each of which is conducted on a separate channel portions at a different scale. 
Consequently, HiSViT has the following virtues: multi-scale interactions, long-range spatial-temporal modeling, and computational efficiency. 

The contributions of this work are summarized as follows:
\begin{itemize}
\setlength{\itemsep}{0mm}
   \item We first offer an insight on the mixed degradation of video SCI and reveal the resulting information skewness between spatial and temporal dimensions. To this end, we make several reasonable modifications on reconstruction architecture and Transformer block.
   \item We propose an efficient video Transformer, dubbed HiSViT, in which CSS-MSA captures long-range cross-scale spatial-temporal dependencies while tackling the information skewness and GSM-FFN enhances the locality.
    \item Extensive experiments demonstrate that our model achieves SOTA performance with comparable or fewer complexity and parameters (see \cref{fig:performance}).
\end{itemize}

\section{Related Work}
\label{sec:related_work}
{\bf Video SCI Reconstruction.}
In recent years, deep learning approaches have extensively exploited on straight~\cite{Cheng2021_CVPR_ReverSCI,Wang2021_CVPR_MetaSCI,wang2022spatial,wang2023efficientsci}, U-shaped~\cite{qiao2020deep,wang2023deep}, recurrent~\cite{Cheng20ECCV_Birnat,cheng2022recurrent}, unrolling~\cite{Ma19ICCV,wu2021dense,yang2022ensemble,meng2023deep,Zheng_2023_ICCV}, and plug-and-play~\cite{Yuan2020_CVPR_PnP,yuan2021plug} architectures with significant performance gains over traditional optimization algorithms~\cite{yuan2014low,Yuan16ICIP_GAP,Liu19_PAMI_DeSCI}.
CNN-based models are impeded by the limited perception field and static kernel of convolution.
Recently, Transformer-based models~\cite{wang2022spatial,wang2023efficientsci,Zheng_2023_ICCV} have achieved SOTA performance.
STFormer~\cite{wang2022spatial} captures spatial and temporal dependencies in parallel and separately with the combination of factorized attention~\cite{bertasius2021space,arnab2021vivit} and windowed attention~\cite{liu2021swin}.
By replacing spatial windowed attention of STFormer with 2D convolutions, EfficientSCI~\cite{wang2023efficientsci} further improve the performance with less computational loads.
CTM-SCI~\cite{Zheng_2023_ICCV} first applies 3D windowed attention~\cite{liu2022video} in video space to enjoy the joint spatial-temporal modelling but its performance gain is at the cost of extremely high complexity and parameters.
By re-examining previous works, we observe that long-range spatial-temporal modeling is desired but the resulting high complexity is troublesome.

{\noindent\bf Vision Transformers.}
Transformers~\cite{vaswani2017attention} have exhibited extraordinary performance on natural language processing tasks~\cite{kenton2019bert,brown2020language} and computer vision tasks~\cite{carion2020end,dosovitskiy2020image,chen2021pre,wang2022uformer}.
As a core of Transformer, vanilla attention suffers from the quadratic computational complexity towards token number and thus is impractical for large-scale dense prediction tasks.
To this end, kinds of Transformer variants~\cite{wang2021pyramid,liu2021swin,zhudeformable,lu2021soft,wang2022kvt,zamir2022restormer} are proposed to decrease the complexity, among which Swin Transformer~\cite{liu2021swin,liu2022video} achieves a good trade-off between accuracy and efficiency by limiting attention calculations within local windows.
Benefitting from long-range dependency and data dependency~\cite{park2022vision},
Transformers have become the de-facto standard of image restoration tasks~\cite{chen2022cross,liang2021swinir,zhangaccurate,mei2023pyramid,chen2024recursive,qu2024dual}.
Due to an additional temporal dimension, developing Transformer for video is more challenging.
Existing video Transformers generally apply spatial (2D) attention under recurrent architecture~\cite{liang2024vrt,liang2022recurrent}, joint spatial-temporal (3D) attention within local windows~\cite{liu2022video}, or factorized spatial-temporal (2D$+$1D) attention~\cite{arnab2021vivit,bertasius2021space}.
All of them are workable but lack appropriate inductive biases from the mixed degradation of video SCI. 

\section{Rethinking Video SCI Reconstruction}
\label{sec:rethinking}
\begin{figure*}[th]
   \centering
    \includegraphics[width=0.96\linewidth]{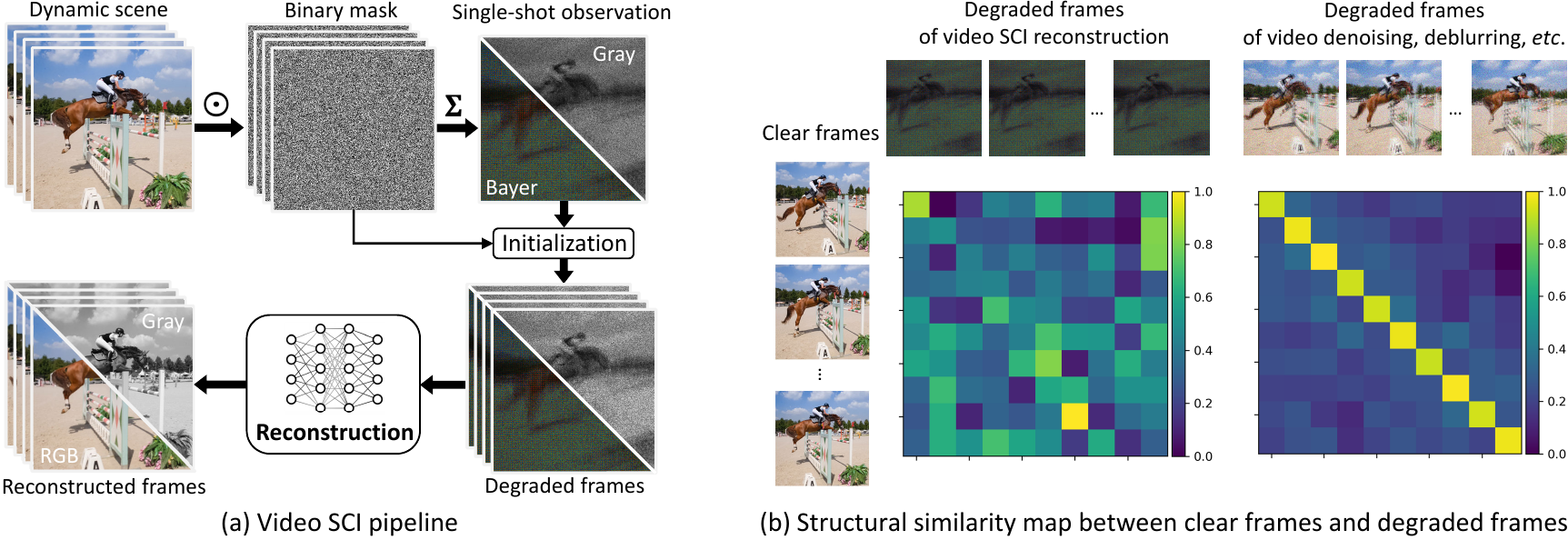}
    \caption{Video SCI pipeline and its degradation. (a) involves the mixed degradation of spatial masking and temporal aliasing, caused by modulation ($\odot$) and multiplexing ($\Sigmamat$). (b) is the structural similarity map between degraded frames and clear frames.}
 \label{fig:pipeline}
 \end{figure*}

\subsection{Mathematical Model}
In video SCI paradigm, a grayscale video $\Vmat \!\in\! \mathbb{R}^{H \!\times\! W \!\times\! T}$ is modulated by mask $\Mmat \!\in\! \mathbb{R}^{H \!\times\! W \!\times\! T}$ and then temporally integrated into an observation $\Imat \!\in\! \mathbb{R}^{H \!\times\! W}$ by
\begin{equation}\label{eq:forward}
\Imat(x,y) \!=\! {\sum\nolimits_{t = 1}^{T} {\Mmat(x,y,t) \!\odot\! \Vmat(x,y,t)} \!+\! \Thetamat(x,y) },
\end{equation}
where $(x,y,t)$ indexes one position in 3D video space, $\odot$ denotes the Hadamard (element-wise) product, and $\Thetamat$ is the measurement noise. Note that color channel is omitted for clarity. For hardware implementation, mask is often generated from a Bernoulli distribution with equal probability, \ie, $\Mmat \!\in\! \{0, 1\} $.

The inverse problem of video SCI is to reconstruct a high-fidelity estimate of $\Vmat$ from the observed $\Imat$.
For dimensional consistency, a highly-degraded video $\bar\Vmat$ is initialized from known $\Imat$ and $\Mmat$ as input by
\begin{equation}
{\bar\Vmat(x,y,t)} \!=\! {\Mmat(x,y,t)} \!\odot\! {\bar \Imat}(x,y), 
~~ {\rm where} ~~
{\bar \Imat}(x,y) \!=\! {\Imat(x,y)} \!\oslash\! {\sum\nolimits_{t = 1}^{T} {\Mmat(x,y,t)}},
\label{eq:intialize}
\end{equation}
where $\oslash$ denotes the element-wise division.
The above 2D-to-3D projection is driven by the pseudoinverse in optimization theory~\cite{boyd2011distributed,liao2014generalized}.
${\bar \Imat} \!\in\! \mathbb{R}^{H \times W}$ is a single-frame coarse estimation of $\Vmat$, whose moving region is blurred and masked but motionless region is closed to the ground truth.
\cref{eq:intialize} implies that ${\bar\Vmat}$ lose the temporal correlations of $\Vmat$ completely (see the inputs in \cref{fig:feature}), whereas imposed by the temporal stamps of $\Mmat$. 
A deep reconstruction model aims to learn a nonlinear map $\mathcal{D}$ from $\bar \Vmat$ to $\Vmat$, namely ${\Vmat} \!=\! \mathcal{D}({\bar \Vmat})$.

\subsection{Degradation Analysis}
For the perspective of imaging in~\cref{eq:forward} and~\cref{fig:pipeline} (a), video SCI involves multiple degradations: spatial masking, temporal aliasing, and measurement noise.
Note that in color video SCI case, demosaicing the observed image cannot recover the right color since optical masks collide with Bayer filter, thus color degradation must be considered.
Among these degradations, the mixture of spatial masking and temporal aliasing is the root of ill-posedness.
\cref{fig:pipeline} (b) visualizes the structural similarity map between clear frames and degraded frames for video SCI reconstruction and a plain video restoration task.
Clearly, the input frames of a plain video restoration task are temporally aligned with clear frames and still contains rich motion dynamics even degraded.
For video SCI reconstruction, the input frames in~\cref{eq:intialize} are the results of re-modulating an identical image ${\bar \Imat}$ with non-semantic different masks ${\Mmat}$ and thus lose temporal correlations (motion dynamics) completely.
As a result, informative clues concentrate spatial dimensions rather than temporal dimension, referred to as {\em information skewness}.

\begin{wrapfigure}{r}{0.5\textwidth}
\centering
\includegraphics[width=0.41\textwidth,trim= 0.5cm 0.4cm 0.5cm 0.2cm]{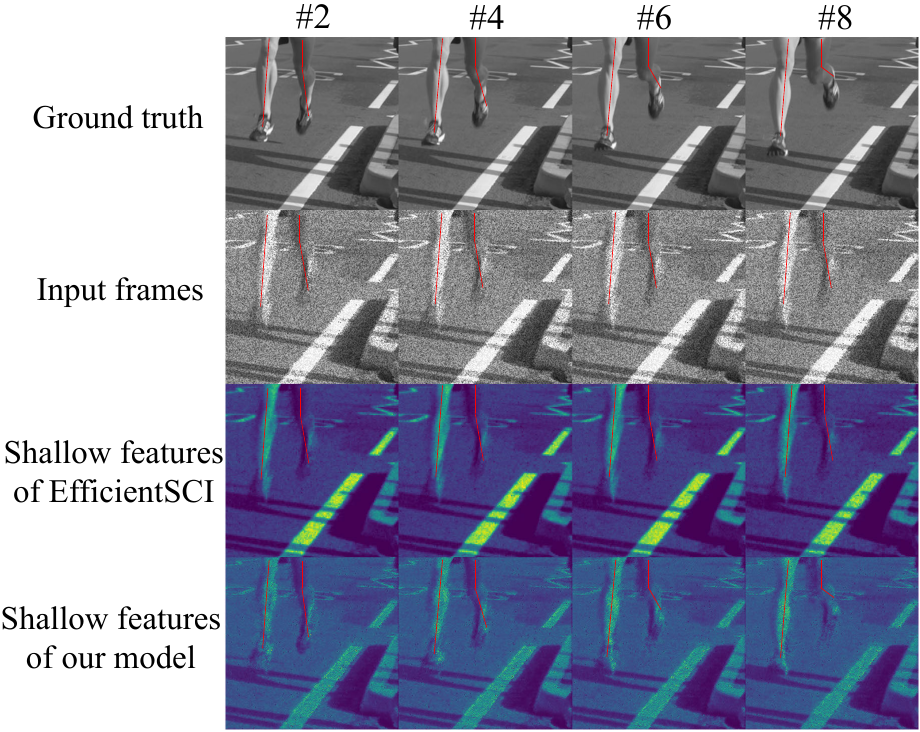}
\caption{Visualization of shallow features extracted by 3D CNN in EfficientSCI~\cite{wang2023efficientsci} and RSTB (without temporal aggregation) in our model. 
Clearly, our frame-wise extraction can better retrieve the temporal correlations with fewer parameters ($0.28$ v.s. $1.12$ M) and MACs ($148.85$ v.s. $241.79$ G).
}
 \label{fig:feature}
\end{wrapfigure}

Unfortunately, previous works have always overlooked the information skewness and follow general vision architectures and blocks, \eg, a recurrent or unrolling architecture with Swin Transformer block.
Due to the information skewness, we observe that too early temporal aggregation is ineffective for temporal dealiasing as demonstrated in~\cref{fig:feature}.
Besides, previous video Transformers, powered by 3D windowed~\cite{liu2022video} or 2D$+$1D factorized~\cite{arnab2021vivit,bertasius2021space} attention, lack an appropriate inductive bias to harmonize with the information skewness.
To this end, we tailor an efficient reconstruction architecture and Transformer block for video SCI reconstruction.

\section{Methodology}
\label{sec:method}
\subsection{Video SCI Reconstruction Architecture}

\cref{fig:baseline} depicts the proposed reconstruction architecture, mainly composed of $i)$ frame-wise feature extraction, $ii)$ spatial-temporal feature refinement, and $iii)$ feature-to-frame reconstruction.
Considering that the feature refinement module generally needs extensive calculations, we propose the downsampling-refinement-upsampling pipeline to relieve computational loads.
For the downsample layer, we use a $1\!\times\!3\!\times\!3$ convolution with a stride of $1\!\times\!2\!\times\!2$ followed by a non-linear activation to decrease the spatial resolution while increasing channels.
For the upsample layer, we use the pixel-shuffle operator to recover the spatial resolution.

\subsubsection{Frame-wise Feature Extraction.}
Due to the loss of motion dynamics caused by the mixed degradation of video SCI, temporal interactions in early layers could exaggerate artifacts owing to error accumulation during propagation~\cite{chan2022investigating}.
With this insight, we first consider the input frames as individual images to process in parallel within the feature extraction module.
Inspired by the effectiveness of using Swin Transformer as the feature extractor~\cite{liang2022recurrent}, we use one Residual Swin Transformer Block (RSTB)~\cite{liang2021swinir} to replace stacked 3D convolutions widely used in previous SOTA models~\cite{Cheng2021_CVPR_ReverSCI,wu2021dense,wang2022spatial,wang2023efficientsci,Zheng_2023_ICCV}.
As visualized in~\cref{fig:feature}, such displacement is more effective and efficient for temporal dealiasing. A clear performance gain is got from \cref{tab:ablation1} in ablation study.
Note that temporal correlations can only be roughly retrieved for simple dynamic scenes. Fine temporal dealiasing relies on the following module.

\subsubsection{Spatial-Temporal Feature Refinement.}
The feature refinement module is built by stacked building blocks to refine the downsampled shallow features from the frame-wise feature extraction module.
In this work, the building block is an efficient Hierarchical Separable Video Transformer (HiSViT), followed by a channel attention~\cite{hu2018squeeze}. HiSViT is detailedly introduced in \cref{subsec:hisvit}.

\subsubsection{Feature-to-Frame Reconstruction.}
The reconstruction module is responsible for generating high-fidelity video frames from the upsampled refined features and shallow features.
Spatial-temporal aggregation or spatial-only aggregation is feasible for this module.
Considering the sufficient spatial-temporal modeling of HiSViT, we use another RSTB for effective reconstruction.

\begin{figure*}[t]
   \centering
    \includegraphics[width=0.955\linewidth]{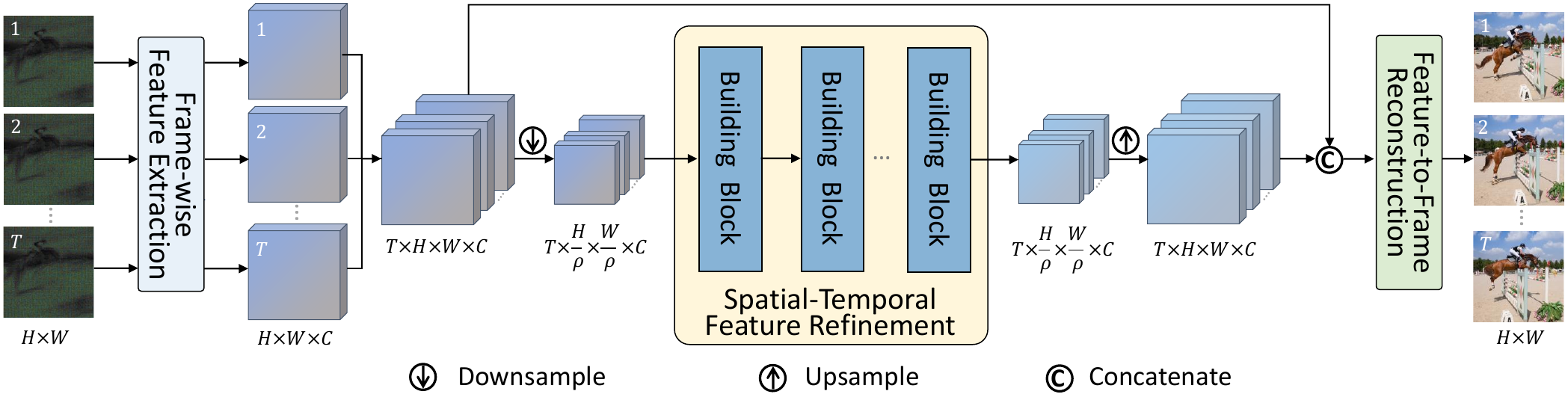}
    \caption{Illustration of the proposed video SCI reconstruction architecture.
    }
 \label{fig:baseline}
 \end{figure*}

\subsection{Hierarchical Separable Video Transformer}
\label{subsec:hisvit}

\begin{figure*}[t]
   \centering
    \includegraphics[width=1\linewidth]{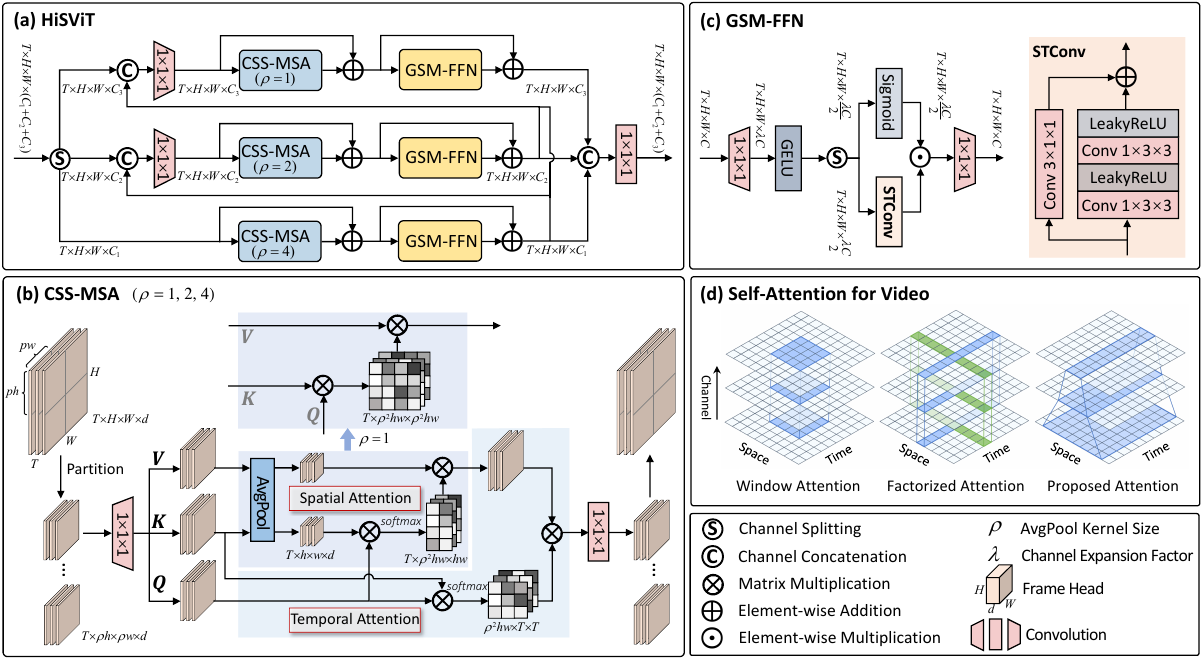}
    \caption{Illustration of HiSViT. (a) the input is split into several portions along channel and then fed into different branches with dense connections. Each branch is composed of a residual (b) CSS-MSA and (c) GSM-FFN.
    Unlike previous windowed or factorized attention, CSS-MSA has an varying receptive field along channel (d). }
 \label{fig:hisvit}
 \end{figure*}

To harmonize with the information skewness of video SCI, we propose a Hierarchical Separable Video Transformer (HiSViT) as building block for efficient spatial-temporal modeling. As demonstrated in \cref{fig:hisvit} (a), HisViT is a multi-branch structure with dense connections along channel dimension and each branch involves a residual Cross-Scale Multi-head Self-Attention (CSS-MSA) and Gated Self-Modulated Feed-Forward Network (GSM-FFN).
CSS-MSA, a spatial-then-temporal attention, attends to all features (tokens) within local windows (across time) and the spatial attention is conducted between normal query and average-pooled key and value, where $\rho$ is the size of spatial average-pooling.
GSM-FFN further strengthens the locality.
As a result, HiSViT has a hierarchical receptive field from bottom ($\rho\!=\!4$) to top ($\rho\!=\!1$) to enable multi-scale interactions and long-range dependencies.

\subsubsection{Cross-Scale Separable Multi-head Self-Attention.}
As shown in \cref{fig:hisvit} (b), CSS-MSA is powered by: $i)$ separating spatial operations from temporal ones; $ii)$ performing cross-scale spatial attention. Inspired by separable convolutions, we decompose regular attention~\cite{dosovitskiy2020image,liu2022video}, which requires intensive interactions in 3D space, into a spatial widowed attention followed by a temporal global attention within a single attention layer.
Spatial attention is conducted between normal query and average-pooled key and value to capture different-frequency information ($\rho\!=\!1,2,4$) given that averaging is a low-pass filter~\cite{voigtman1986low}.

Here to simplify the presentation, we describe only a single head of CSS-MSA.
At a certain branch with $\rho$, let $\Xmat \!\in\! \mathbb{R}^{T\!\times\! H\!\times\! W\!\times\! d}$ be the input video feature.
$\Xmat$ are first partitioned into several non-overlapped patches $\Xmat_{i}\!\in\! \mathbb{R}^{T\!\times\! \rho h\!\times\! \rho w\!\times\! d}, i\!=\!1,..., {\frac{{HW}}{{{\rho ^{\rm{2}}}hw}}}$ according to spatial window $\rho h \!\times\! \rho w$. 
Afterwards, query $\Qmat_{i}$, key $\Kmat_{i}$, and value $\Vmat_{i}$ are computed from $\Xmat_{i}$ by
\begin{equation}
\begin{array}{c}
{\Qmat_{i} \!=\! \Xmat_{i} \Wmat^{q},
\qquad
\Kmat_{i} \!=\! \Xmat_{i} \Wmat^{k}, 
\qquad 
\Vmat_{i} \!=\! \Xmat_{i} \Wmat^{v},}
\end{array}
\label{eq:qkv}
\end{equation}
where $\Qmat_{i}, \Kmat_{i}, \Vmat_{i} \!\in\! \mathbb{R}^{T\!\times\! ph\!\times\! pw\!\times\! d}$ and $\Wmat^{\{q,k,v\}} \!\in\! \mathbb{R}^{d\!\times\! d}$ represent learnable projection matrices.
If $\rho \!> \!1$, $\Kmat_{i}$, $\Vmat_{i}$ are spatially average-pooled into ${\Kmat}_{i}^{\downarrow}, {\Vmat}_{i}^{\downarrow} \!\in\! \mathbb{R}^{T\!\times\! h\!\times\! w\!\times\! d}$, otherwise there is no pooling operator, \ie, $\Kmat_{i} \!=\! {\Kmat}_{i}^{\downarrow}, {\Vmat}_{i}\!=\!{\Vmat}_{i}^{\downarrow}$.
CSS-MSA aggregates spatial-temporal features using ${\Qmat}_{i}$, ${\Kmat}_{i}$, ${\Kmat}_{i}^{\downarrow}$, and ${\Vmat}_{i}^{\downarrow}$ by
\begin{align}
\begin{split}
\Vmat'_{i} \!=\! \mathtt{softmax} ({{{\Qmat}_{i} {{\Kmat}_{i}^{\downarrow\top}}} \mathord{\left/
{\vphantom {{{\Qmat}_{i} {{\Kmat}_{i}^{\downarrow\top}}} {\tau_1 }}} \right.
\kern-\nulldelimiterspace} {\tau_1 }}){{\Vmat}_{i}^{\downarrow}}&, \\  
{\rm where} \quad {\Qmat}_{i} \!\in\!  \mathbb{R}^{T \!\times\! {{\rho ^{\rm{2}}}hw} \!\times\! d} \!\gets\! \mathbb{R}^{T\!\times\! ph\!\times\! pw\!\times\! d},
~~~{\Kmat}_{i}^{\downarrow}, {\Vmat}_{i}^{\downarrow}& \!\in\! \mathbb{R}^{T \!\times\! {hw} \!\times\! d} \!\gets\! \mathbb{R}^{T \!\times\! h \!\times\! w \!\times\! d},
\end{split}
\\
\begin{split}
\Vmat''_{i} \!=\! \mathtt{softmax} ({{{\Qmat}_{i} {{\Kmat}_{i}^{\top}}} \mathord{\left/
{\vphantom {{{\Qmat}_{i} {{\Kmat}_{i}^{\top}}} {\tau_2 }}} \right.
\kern-\nulldelimiterspace} {\tau_2 }}){\Vmat'_{i}}&, \\  
{\rm where} \quad {\Qmat}_{i},{\Kmat}_{i} \!\in\! \mathbb{R}^{{{\rho ^{\rm{2}}}hw} \!\times\! T\!\times\! d} \!\gets\! \mathbb{R}^{T\!\times\! ph\!\times\! pw\!\times\! d},
~\Vmat'_{i} \!\in &\mathbb{R}^{{{\rho ^{\rm{2}}}hw} \!\times\! T \!\times\! d} \!\gets\! \mathbb{R}^{T \!\times\! {{\rho ^{\rm{2}}}hw} \!\times\! d},
\end{split}
\label{eq:temporal_attention}
\end{align}
Note that the above matrix multiplications are batch-wise and ${\tau_1}, {\tau_2}$ are two learnable scales.
The output is computed by a linear projection $\Ymat_{i} \!=\! \Vmat''_{i}\Wmat$, where $\Vmat''_{i} \!\in\! \mathbb{R}^{T \!\times\! \rho h \!\times\! \rho w \!\times\! d} \gets \mathbb{R}^{{{\rho ^{\rm{2}}}hw} \!\times\! T\!\times\! d}$ and $\Wmat \!\in\! \mathbb{R}^{d \!\times\! d}$ is learnable. $\{\Ymat_{i}\}_{i\!=\!1}^{N} \!\in\! \mathbb{R}^{T \!\times\! \rho h \!\times\! \rho w \!\times\! d}$ ($N\!=\!{\frac{{HW}}{{{\rho ^{\rm{2}}}hw}}}$) are combined into the final output $\Ymat \!\in\! \mathbb{R}^{T \!\times\! H \!\times\! W \!\times\! d}$.
Clearly, the input and output have the same size regardless of the pooling size $\rho$. We adapt the shifted rectangle-window strategy~\cite{chen2022cross} for spatial partition.

\begin{figure}[t]
\begin{minipage}[c]{.48\linewidth}
\centering
\scalebox{0.75}{
\begin{tabular}{c|c}
\toprule[0.15em]
Method & Computational Complexity \\
\midrule[0.1em]
{$\Omega$(G-MSA)~\cite{dosovitskiy2020image}} &
$4THW{d^2} \!+\! 2{(THW)^2}d$\\
$\Omega$(W-MSA)~\cite{liu2022video} & 
$4THW{d^2} \!+\! 2thwTHWd$ \\
$\Omega$(F-MSA)~\cite{arnab2021vivit,bertasius2021space} & 
$8THW{d^2} \!+\! 2{T^2}HWd \!+\! 2T{(HW)^2}d$\\
$\Omega$(SW-MSA) & $4THW{d^2} \!+\! 2hw{T^2}HWd$ \\
$\Omega$(FW-MSA) & $8THW{d^2} \!+\! 2{T^2}HWd \!+\! 2hwT{HW}d$\\
\midrule
$\Omega$(CSS-MSA) & $4THW{d^2} \!+\! 2{T^2}HWd \!+\! 2hwT\frac{HW}{\rho^2}d$
\\
\bottomrule[0.1em]
\end{tabular}
}
\captionof{table}{Computational complexity of different MSAs for an input size {\small $T\!\times\! H\!\times\!W\!\times\!d$}. {\small $t\!\times\! h\!\times\!w$} denotes the 3D window size. $\rho$ is the spatial average pooling size.}
\label{tab:complexity}
\end{minipage}
\begin{minipage}[c]{.5\linewidth}
\centering
\includegraphics[width=6cm, trim= 0cm 0.35cm 0cm 0cm]{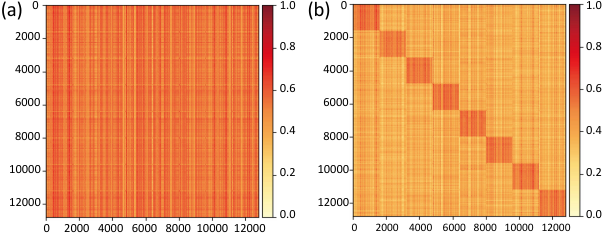}
\caption{Visualization of attention matrix from $8\!\times\!40\!\times\!40$ tokens. (a) is directly computed from query and key. (b) is an equivalent fusion of spatial and temporal attention matrices on CSS-MSA ($\rho\!= \!1$).}
\label{fig:attention_map}
\end{minipage}
\end{figure}

\subsubsection{Comparison with Mainstream MSAs.}
Essentially, CSS-MSA is a spatial-then-temporal attention, namely a spatial windowed attention followed by a temporal global attention within a single attention layer.
Next, we compare it with mainstream attention mechanisms for video, including G-MSA~\cite{dosovitskiy2020image}, W-MSA~\cite{liu2022video}, F-MSA~\cite{bertasius2021space,arnab2021vivit}, and their variants. The computational complexity is summarized in \cref{tab:complexity}. G-MSA and F-MSA suffer from the quadratic computational complexity towards spatial-temporal resolution $T\!\times\! H\!\times\!W$ and spatial resolution $H\!\times\!W$ respectively. W-MSA has the linear complexity at the cost of limiting interactions within $t\!\times\! h\!\times\!w$ local windows. 
For long-range temporal dependencies, an variant is to relax 3D window $t\!\times\! h\!\times\!w$ into 2D window $ h\!\times\!w$ for video, referred to as Spatially-Windowed MSA (SW-MSA).
A hybrid of F-MSA and W-MSA is to perform spatial windowed MSA and temporal global MSA in two separate attention layers, referred to as FW-MSA.
Unlike FW-MSA, CSS-MSA attends to all spatial-temporal tokens with cross-scale interactions in a single attention layer and is equivalent to a joint spatial-temporal attention matrix in~\cref{fig:attention_map} (b). Compared with regular attention in~\cref{fig:attention_map} (a), the proposed CSS-MSA pays more attention to intraframe rather than interframe aggregation while keeping long-range spatial-temporal modeling ability.
A quantitative comparison is given in \cref{tab:ablation3} of ablation study.

\subsubsection{Gated Self-Modulated Feed-Forward Network.}
As another key component, regular FFN process the output from MSA layer with a simple residual structure, built by two linear projections and a nonlinear activation between them.
Here, we propose GSM-FFN by making two fundamental modifications on FFN: $i)$ Gated Self-Modulation (GSM) and $ii)$ factorized Spatial-Temporal Convolution (STConv).
As depicted in \cref{fig:hisvit} (c), given the input feature $\Xmat \!\in\! \mathbb{R}^{T\!\times\! H\!\times\! W\!\times\! C}$ from CSS-MSA layer, the output feature is computed by
\begin{equation}
\begin{array}{c}
{\Xmat_1, \Xmat_2 \!=\! \mathtt{Split}(\mathtt{GELU}(\Xmat{\Wmat_1}))}\\
{\Ymat \!=\! (\mathtt{Sigmoid}(\Xmat_1) \!\odot\! \mathtt{STConv}(\Xmat_2)){\Wmat_2}},
\end{array}
\label{eq:ffn}
\end{equation}
where $\Wmat_1 \!\in\! \mathbb{R}^{ C\!\times\! \lambda C}$ increases the channel number by $\lambda$ times, $\Wmat_2 \!\in\! \mathbb{R}^{\frac{{\lambda C}}{{\rm{2}}}\!\times\! C}$ regulates the channel number into $C$, $\mathtt{Split}$ divides the channels into half, $\mathtt{GELU}$ is a non-linear activation function, and $\mathtt{Sigmoid}$ represent the sigmoid function. Inspired by~\cite{lai2023hybrid,chen2023dual}, $\mathtt{STConv}$, a hybrid of 1D convolution, 2D convolution, and LeakyReLU, performs convolutional and non-linear operators in spatial and temporal dimensions separately as shown in the right of \cref{fig:hisvit} (c).

\section{Experiments}
\label{sec:experiment}

\subsubsection{Model Setting.} We use HiSViT in~\cref{fig:hisvit} as building block of the proposed reconstruction architecture in \cref{fig:baseline}.
In the frame-wise feature extraction and feature-to-frame reconstruction modules, the channel number of RSTB~\cite{liang2021swinir} is set to $128$.
In the feature refinement module, the channel number of three branches is set to $64,64,128$ for $p\!=\!1,2,4$ and the channel expansion factor of GSM-FFN is set to $\lambda \!=\!2$.
To explore the scalability of HiSViT, we define two model settings: \textbf{HiSViT\mysubscript{9}} and \textbf{HiSViT\mysubscript{13}}, which involves $9$ and $13$ building blocks respectively.

\subsubsection{Experiment Setting.} To validate the effectiveness of the proposed method, we conduct experiments on six grayscale/color benchmark videos with the resolution of $8\!\times\!256\!\times\!256$/$8\!\times\!512\!\times\!512\!\times\!3$ pixels and on real captured grayscale videos~\cite{qiao2020deep} with the resolution of $10\!\times\!512\!\times\!512$ pixels. 
Following previous works, our models are trained in DAVIS2017 dataset~\cite{pont20172017} with the same data augmentation in~\cite{wang2022spatial,wang2023efficientsci}. We use MSE loss with Adam optimizer (${\beta_1} \!=\! 0.9$, ${\beta_2} \!=\! 0.999$) on A100 GPUs.
All models are pretrained on the resolution of $8\!\times\!128\!\times\!128(\!\times\!3)$ with a $1\!\times\!10^{-4}$ learning ratio over $100$ epochs and then fine-tuned on the resolution of $8\!\times\!256\!\times\!256(\!\times\!3)$ with a $1\!\times\!10^{-5}$ learning ratio over $20$ epochs.
Peak Signal to Noise
Ratio (PSNR) and Structural SIMilarity (SSIM) are used to measure the reconstruction fidelity.
Multiply-ACcumulate operations (MACs) are used to measure the computational complexity.
More model details and additional results are in the supplementary material.
In all experiments, the best and second-best results of the evaluated methods are \textbf{highlighted} and \underline{underlined}.

\begin{figure}[t]
   \centering
    \includegraphics[width=0.9\linewidth]{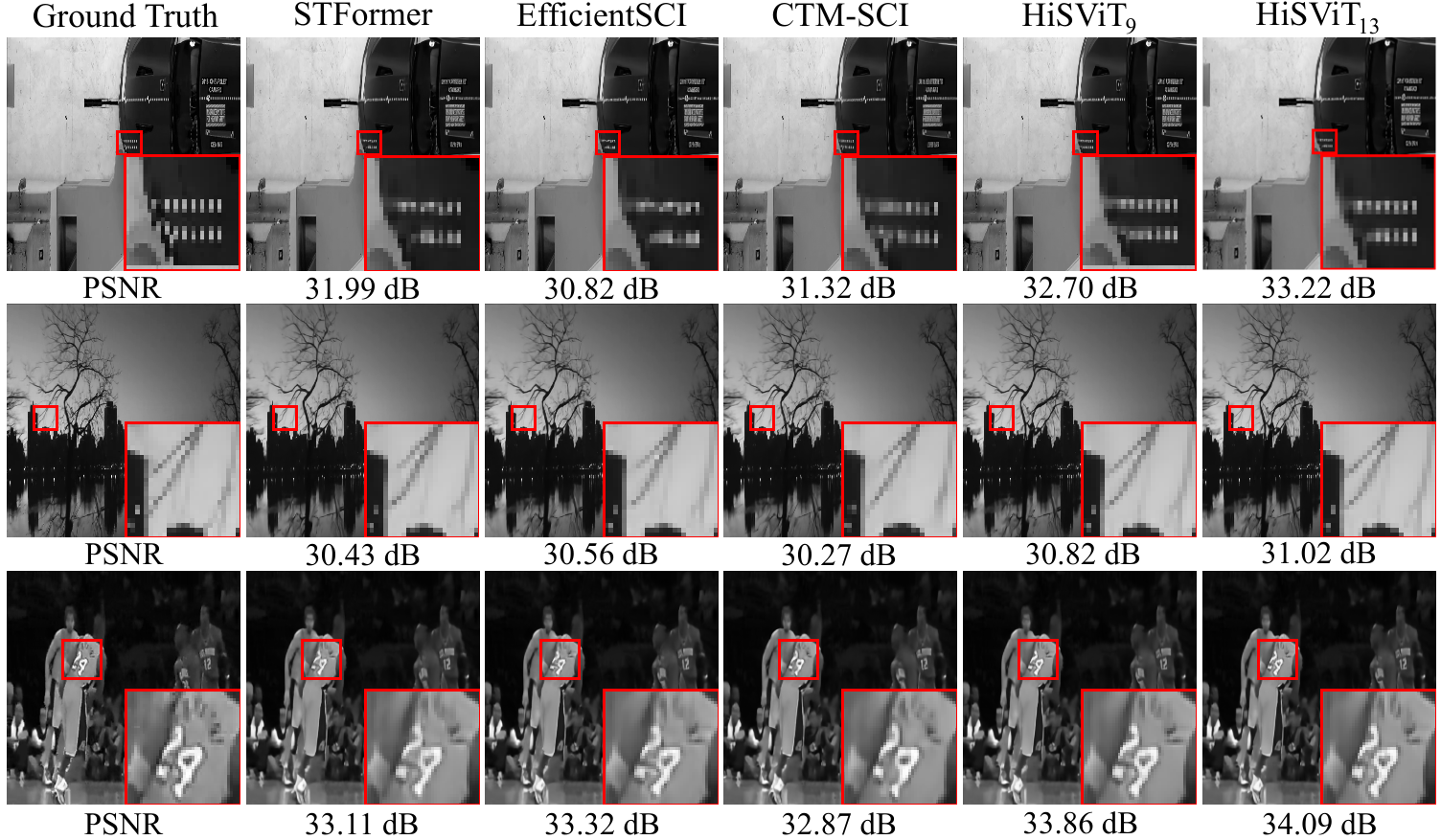}
    \caption{Visual results of competitive methods on grayscale video frames.}
 \label{fig:grayscale_results}
\end{figure}

\begin{table*}[t]
\begin{center}
\caption{Quantitative results of different methods on \underline{\textbf{grayscale videos}} in terms of PSNR (dB)~$\textcolor{black}{\uparrow}$, SSIM~$\textcolor{black}{\uparrow}$, Params (M)~$\textcolor{black}{\downarrow}$, and MACs (G)~$\textcolor{black}{\downarrow}$.}
\label{table:grayscale_results}
\setlength{\tabcolsep}{3pt}
\scalebox{0.56}{
\begin{tabular}{l |c |c |c c c c c c || c }
\toprule[0.15em]
 \textbf{Method} &
   \textbf{Params} &
  \textbf{MACs} &
 \textbf{Kobe} &
 \textbf{Traffic} &  
 \textbf{Runner} & 
 \textbf{Drop} & 
 \textbf{Crash} & 
 \textbf{Aerial} & 
 \textbf{Average} \\
\midrule[0.15em]
GAP-TV~\cite{Yuan16ICIP_GAP} & -- & --& 26.46, 0.885  & 20.89, 0.715  & 28.52, 0.909  & 34.63, 0.970  & 24.82, 0.838 & 25.05, 0.828  & 26.73, 0.858\\
DeSCI~\cite{Liu19_PAMI_DeSCI} & -- & --& 33.25, 0.952  & 28.71, 0.925  & 38.48, 0.969  & 43.10, 0.993  & 27.04, 0.909 & 25.33, 0.860  & 32.65, 0.935\\
PnP-FFDNet~\cite{Yuan2020_CVPR_PnP} & -- & -- &30.50, 0.926& 24.18, 0.828& 32.15, 0.933& 40.70, 0.989& 25.42, 0.849& 25.27, 0.829& 29.70, 0.892\\
PnP-FastDVDnet~\cite{yuan2021plug} & -- & --& 32.73, 0.947  & 27.95, 0.932  & 36.29, 0.962  & 41.82, 0.989  & 27.32, 0.925 & 27.98, 0.897  & 32.35, 0.942\\
E2E-CNN~\cite{qiao2020deep} & 0.82 & 53.48 & 27.79, 0.807 & 24.62, 0.840 & 34.12, 0.947 & 36.56, 0.949 & 26.43, 0.882 & 27.18, 0.869 & 29.45, 0.882\\
BIRNAT~\cite{Cheng20ECCV_Birnat} & 4.13 & 390.56 & 32.71, 0.950  & 29.33, 0.942  & 38.70, 0.976  & 42.28, 0.992 &
27.84, 0.927 & 28.99, 0.917& 33.31, 0.951\\
GAP-net-Unet-S12~\cite{meng2023deep} &  5.62 & 87.38 & 32.09, 0.944  & 28.19, 0.929  & 38.12, 0.975  & 42.02, 0.992 &
27.83, 0.931 & 28.88, 0.914& 32.86, 0.947\\
MeteSCI~\cite{Wang2021_CVPR_MetaSCI} & 2.89 & 39.85 &30.12, 0.907& 26.95, 0.888& 37.02, 0.967& 40.61, 0.985& 27.33, 0.906& 28.31, 0.904& 31.72, 0.926\\
RevSCI~\cite{Cheng2021_CVPR_ReverSCI} & 5.66 & 766.95 & 33.72, 0.957 & 30.02, 0.949 & 39.40, 0.977 & 42.93, \underline{0.992} & 28.12, 0.937 & 29.35, 0.924 & 33.92, 0.956\\
DUN-3DUnet~\cite{wu2021dense} & 61.91 & 3975.83 & 35.00, 0.969 & 31.76, 0.966 & 40.03, 0.980 & 44.96, \textbf{0.995} & 29.33, 0.956 & 30.46, 0.943 & 35.26, 0.968\\
ELP-Unfolding~\cite{yang2022ensemble} & 565.73 & 4634.94 & 34.41, 0.966 & 31.58, 0.962 & 41,16, 0.986 & 44.99, \textbf{0.995} & 29.65, 0.959 & 30.68, 0.944 & 35.41, 0.969\\
STFormer~\cite{wang2022spatial} & 19.48 & 3060.75 & 35.53, 0.973 & 32.15, 0.967 & 42.64, \underline{0.988} & 45.08, \textbf{0.995} & 31.06, 0.970 & 31.56, 0.953 & 36.34, 0.974\\
EfficientSCI~\cite{wang2023efficientsci} & {8.82} & {1426.38} & 35.76, 0.974 & 32.30, 0.968 & 43.05, \underline{0.988} & 45.18, \textbf{0.995} & 31.13, 0.971 & 31.50, 0.953 & 36.48, 0.975\\
CTM-SCI~\cite{Zheng_2023_ICCV} & 81.81 & 12793.93 & 
35.97, 0.975 & 
32.59, 0.970 & 
42.10, 0.987 & 
45.49, \textbf{0.995} & 
31.33, 0.971 & 
31.64, 0.955 & 
36.52, 0.976\\
\midrule
\textbf{HiSViT}~$\!_{9}$ & {8.98} & {1535.92}
& \underline{36.24}, \underline{0.976}  & \underline{33.06}, \underline{0.973}  & \underline{43.84}, \textbf{0.991}  & \underline{45.55}, \textbf{0.995}  & \underline{31.62}, \underline{0.976} & \underline{31.67}, \underline{0.957}  & \underline{37.00}, \underline{0.978}\\
\textbf{HiSViT}~$\!_{13}$ & 12.16 &  1947.30
& \textbf{36.50}, \textbf{0.979}  & \textbf{33.42}, \textbf{0.975}  & \textbf{44.32}, \textbf{0.991}  & \textbf{45.62}, \textbf{0.995}  & \textbf{31.93}, \textbf{0.978} & \textbf{31.94}, \textbf{0.959}  & \textbf{37.29}, \textbf{0.980}\\
\bottomrule[0.1em]
\end{tabular}
}
\end{center}
\end{table*}

\subsection{Results on Grayscale Benchmark Videos}

We compare HiSViT\mysubscript{9}/HiSViT\mysubscript{13} with two representative optimization algorithms (GAP-TV~\cite{Yuan16ICIP_GAP}, DeSCI~\cite{Liu19_PAMI_DeSCI}), two plug-and-play methods (PnP-FFDNet~\cite{Yuan2020_CVPR_PnP}, PnP-FastDVDnet~\cite{yuan2021plug}), seven CNN-based methods (E2E-CNN~\cite{qiao2020deep}, BIRNAT~\cite{Cheng20ECCV_Birnat}, GAP-net-Unet-S12~\cite{meng2023deep}, MeteSCI~\cite{Wang2021_CVPR_MetaSCI}, RevSCI~\cite{Cheng2021_CVPR_ReverSCI}, DUN-3DUnet~\cite{wu2021dense}, ELP-Unfolding~\cite{yang2022ensemble}), three Transformer-based methods (STFormer~\cite{wang2022spatial}, EfficientSCI~\cite{wang2023efficientsci}, CTM-SCI~\cite{Zheng_2023_ICCV}).
\cref{table:grayscale_results} reports the fidelity scores of all methods, and the parameters and MACs of deep learning methods on six grayscale benchmark videos. 
In terms of the reconstruction fidelity, the proposed HiSViT\mysubscript{9} and HiSViT\mysubscript{13} outperform previous optimization, plug-and-play, and CNN-based methods by a large margin ($>$\textbf{1.5} dB).
Compared with EfficientSCI, our HiSViT\mysubscript{9} outperforms it by \textbf{0.52} dB with comparable parameters and MACs.
Compared with previous best CTM-SCI, our HiSViT\mysubscript{9}/HiSViT\mysubscript{13} outperforms it by \textbf{0.48}/\textbf{0.77} dB with only \textbf{12.00}/\textbf{15.22}~${\%}$ of its MACs and \textbf{10.98}/\textbf{14.86}~${\%}$ of its paramters.
Clearly, our method achieves not only SOTA results but also a good trade-off between performance and efficiency.
\cref{fig:grayscale_results} shows the visual comparison with competitive methods.
Our models can retrieve more details and textures.

\subsection{Results on Color Benchmark Videos}

\begin{figure}[t]
   \centering
    \includegraphics[width=0.91\linewidth]{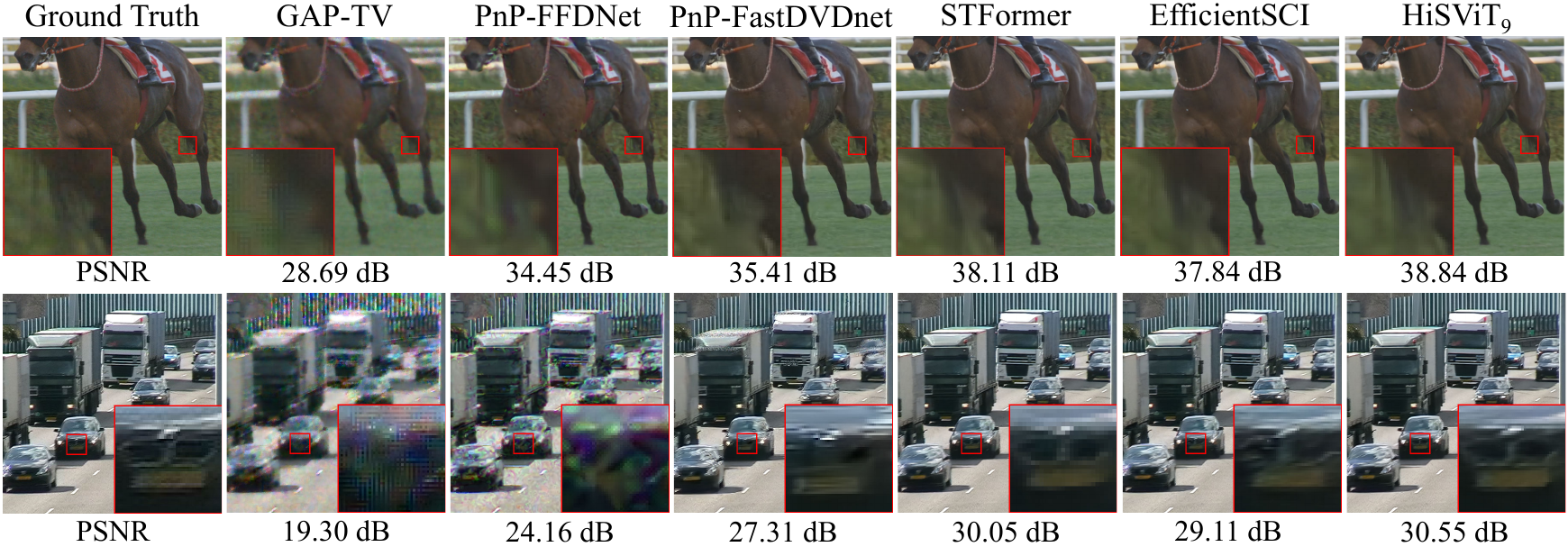}
    \caption{Visual results of competitive methods on color video frames.}
 \label{fig:color_results}
 \end{figure}

\begin{table*}[!tp]
\begin{center}
\caption{Quantitative results of different methods on \underline{\textbf{color videos}} in terms of PSNR (dB)~$\textcolor{black}{\uparrow}$, SSIM~$\textcolor{black}{\uparrow}$, Params (M)~$\textcolor{black}{\downarrow}$, and MACs (G)~$\textcolor{black}{\downarrow}$.}
\label{tab:color_results}
\setlength{\tabcolsep}{3pt}
\scalebox{0.56}{
\begin{tabular}{l |c |c |c c c c c c || c }
\toprule[0.15em]
 \textbf{Method} &
 \textbf{Params} & 
 \textbf{MACs} & 
 \textbf{Beauty} &
 \textbf{Bosphorus} &  
 \textbf{Jockey} & 
 \textbf{Runner} & 
 \textbf{ShakeNDry} & 
 \textbf{Traffic} & 
 \textbf{Average} \\
\midrule[0.15em]
GAP-TV~\cite{Yuan16ICIP_GAP} & --& --& 33.08, 0.964& 29.70, 0.914& 29.48, 0.887& 29.10, 0.878& 29.59, 0.893& 19.84, 0.645& 28.47, 0.864\\
DeSCI~\cite{Liu19_PAMI_DeSCI} & --&-- & 34.66, 0.971& 32.88, 0.952& 34.14, 0.938& 36.16, 0.949& 30.94, 0.905& 24.62, 0.839& 32.23, 0.926\\
PnP-FFDNet~\cite{Yuan2020_CVPR_PnP} & --&-- &34.15, 0.967& 33.06, 0.957& 34.80, 0.943& 35.32, 0.940& 32.37, 0.940& 24.55, 0.837& 32.38, 0.931\\
PnP-FastDVDnet~\cite{yuan2021plug} & -- &-- & 35.27, 0.972& 37.24, 0.971& 35.63, 0.950& 38.22, 0.965& 33.71, 0.949& 27.49, 0.915& 34.60, 0.953\\
BIRNAT~\cite{Cheng20ECCV_Birnat} & 4.14 & 1454.96 & 36.08, 0.975& 38.30, 0.982& 36.51, 0.956& 39.65, 0.973& 34.26, 0.951& 28.03, 0.915& 35.47, 0.959\\
STFormer~\cite{wang2022spatial} & 19.49 & 12155.47 & 37.37, \underline{0.981}& 40.39, \underline{0.988}& 38.32, 0.968& 42.45, \underline{0.985}& 35.15, \underline{0.956}& \underline{30.24}, 0.939& 37.32, \underline{0.970}\\
EfficientSCI~\cite{wang2023efficientsci}& 8.83 & 5701.50 & \underline{37.51}, {0.979}& \underline{40.89}, \underline{0.988}& \underline{38.49}, \underline{0.969}& \underline{42.73}, \underline{0.985}& \underline{35.19}, {0.953}&  {30.13}, \underline{0.943}& \underline{37.49}, \underline{0.970}\\
\midrule
\textbf{HiSViT}~$\!_{9}$ & 8.98 & 6143.68
& \textbf{37.75}, \textbf{0.981}  & \textbf{41.50}, \textbf{0.990}  & \textbf{39.29}, \textbf{0.972}  & \textbf{43.27}, \textbf{0.986}  & \textbf{35.49}, \textbf{0.958} & \textbf{30.76}, \textbf{0.946}  & \textbf{38.01}, \textbf{0.972}\\
\bottomrule[0.1em]
\end{tabular}}
\end{center}
\end{table*}

As mentioned previously, color video SCI reconstruction must be bound to demosaicing since spatial masking collides with Bayer filter, thus it is a more challenging task than grayscale one. Unfortunately, less effort has been spent on it. Without any specialized designs, we just change the output channel number from $1$ to $3$ for the hybrid task of color video SCI reconstruction and demosaicing. \cref{tab:color_results} reports the quantitative results of available methods (GAP-TV~\cite{Yuan16ICIP_GAP}, DeSCI~\cite{Liu19_PAMI_DeSCI}, PnP-FFDNet~\cite{Yuan2020_CVPR_PnP}, PnP-FastDVDnet~\cite{yuan2021plug}, BIRNAT~\cite{Cheng20ECCV_Birnat}, STFormer~\cite{wang2022spatial}, EfficientSCI~\cite{wang2023efficientsci}).
Our HiSViT\mysubscript{9} outperforms previous best EfficientSCI by \textbf{0.52} dB with comparable parameters and MACs.
\cref{fig:grayscale_results} shows the visual comparison with competitive methods.
Our model is better than previous methods in restoring correct colors and fine structures.

\subsection{Results on Real Captured Videos}
\begin{figure}[!pt]
   \centering
    \includegraphics[width=0.93\linewidth]{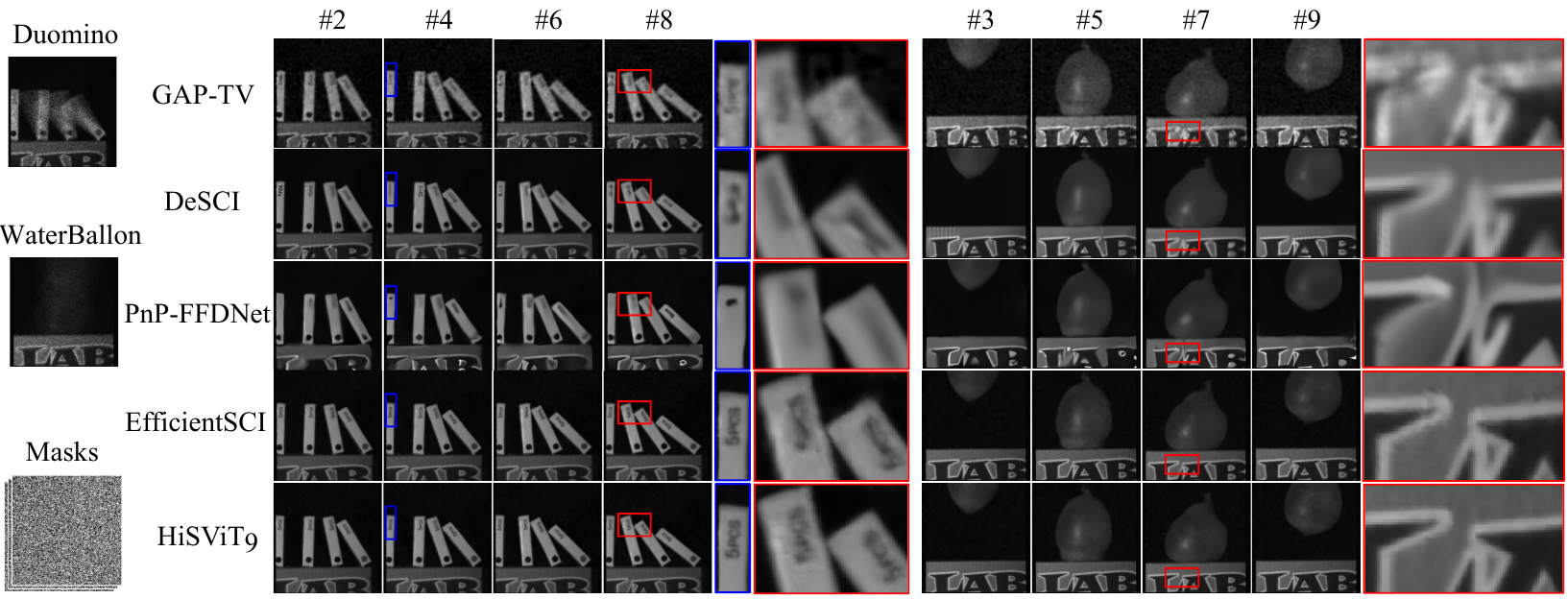}
    \caption{Reconstructed results of real captured {\tt Duonimo} and {\tt WaterBallon}.}
 \label{fig:real_data}
 \end{figure}

We further evaluate our method on two public real observations ({\tt Duonimo} and {\tt WaterBallon}~\cite{qiao2020deep}). For fair competition with EfficientSCI~\cite{wang2023efficientsci}, we use HiSViT\mysubscript{9} to conduct real data testing.
\cref{fig:real_data} shows the visual results reconstructed by GAP-TV~\cite{Yuan16ICIP_GAP}, DeSCI~\cite{Liu19_PAMI_DeSCI}, PnP-FFDNet~\cite{Yuan2020_CVPR_PnP}, EfficientSCI~\cite{wang2023efficientsci}, and our HiSViT\mysubscript{9}.
Clearly, GAP-TV suffers from strong artifacts. DeSCI and PnP-FFDNet lead to over-smoothing results. Transformer-based EfficientSCI and HiSViT\mysubscript{9} show an excellent generalization ability against noises on physical systems and significantly outperform non-Transformer methods.
Compared to EfficientSCI, our HiSViT\mysubscript{9} can better reconstruct image details in the captured scene and avoid the artifacts which are out of the captured scene.

\begin{table*}[t]
\parbox{.5\textwidth}{
\centering
\captionof{table}{Effect of architectural designs.} 
\label{tab:ablation1}
\setlength{\tabcolsep}{1.5pt}
\scalebox{0.7}{
\begin{tabular}{l | c | c | c | c}
\toprule[0.15em]
\textbf{Architecture} & \textbf{Params} &  \textbf{MACs} & \textbf{PSNR} & \textbf{SSIM}   \\
\midrule[0.15 em]
(a) 3D-CNN$+$3D-CNN & 8.86 & 1421.70 & 36.65 & 0.976  \\
(b) 2D-CNN$+$3D-CNN & 9.76 & 1339.58 & 36.79 & 0.976  \\
(c) RSTB$+$3D-CNN & 8.85 & 1466.19 & 36.86 & 0.977  \\
(d) RSTB$+$RSTB  & 8.83 & 1458.61 & 36.92 & 0.978 \\
(e) RSTB$+$RSTB$+$Skip  & 8.98 & 1535.92 & 37.00 & 0.978 \\

\bottomrule[0.1em]
\end{tabular}
}
}
\hfill
\parbox{.5\textwidth}{
\centering
\captionof{table}{Effect of separable and cross-scale designs in CSS-MSA.} 
\label{tab:ablation2}
\setlength{\tabcolsep}{1pt}
\scalebox{0.75}{
\begin{tabular}{c | c | c | c | c | c}
\toprule[0.15em]
\textbf{Cross-scale} & \textbf{Separable} & \textbf{Params} & \textbf{MACs} & \textbf{PSNR} & \textbf{SSIM}    \\
\midrule[0.15 em]
\checkmark & & 8.89 & 2155.08 & 36.57 & 0.977\\
 & \checkmark  & 8.89 & 1769.78 & 36.92 & 0.978 \\
\checkmark & \checkmark  & 8.89 & 1535.92 & 37.00 & 0.978\\
\bottomrule[0.1em]
\end{tabular}}
}
\end{table*}

\begin{table*}[t]
\parbox{.45\textwidth}{
\centering
\captionof{table}{Comparison of CSS-MSA and competitive MSAs. } 
\label{tab:ablation3}
\setlength{\tabcolsep}{1.5pt}
\scalebox{0.75}{
\begin{tabular}{c | c | c | c | c}
\toprule[0.2em]
\textbf{Method} & \textbf{Params} & \textbf{MACs}  & \textbf{PSNR}   & \textbf{SSIM} \\
\midrule[0.15 em]
SW-MSA & 21.90 & 3586.27 & 35.55 & 0.970 \\
FW-MSA & 23.67 & 3552.44 & 35.38 & 0.968 \\
CSS-MSA & 8.89 & 1535.92 & 37.00 & 0.978 \\
\bottomrule[0.1em]
\end{tabular}}
}
\hfill
\parbox{.55\textwidth}{
\centering
\captionof{table}{Effect of various designs in GSM-FFN.} 
\label{tab:ablation4}
\setlength{\tabcolsep}{2pt}
\scalebox{0.75}{
\begin{tabular}{c | c | c | c | c}
\toprule[0.2em]
\textbf{Method} & \textbf{Params} & \textbf{MACs}  & \textbf{PSNR}   & \textbf{SSIM}  \\
\midrule[0.15 em]
Regular FFN &  5.44  & 1072.06 & 36.28  & 0.974\\
w/o GSM & 8.76 & 1506.93 & 36.82 & 0.977\\
w/o STConV & 10.30 & 1709.87 & 36.86 & 0.977\\
GSM-FFN & 8.89 & 1535.92 & 37.00 & 0.978\\
\bottomrule[0.1em]
\end{tabular}
}
}
\end{table*}

\subsection{Ablation Study}
To offer an insight into the proposed method, we demystify the effect of reconstruction architecture and CSS-MSA and GSM-FFN of HiSViT. In addition, we also compare CSS-MSA with competitive MSAs for video SCI reconstruction. The ablation experiments are conducted on grayscale videos towards HiSViT\mysubscript{9}.

\subsubsection{Improvements in Architecture.}
Previous competitive reconstruction models, including STFormer~\cite{wang2022spatial}, EfficientSCI~\cite{wang2023efficientsci}, CTM-SCI~\cite{Zheng_2023_ICCV}, always use 3D CNN for shallow feature extraction and feature-to-frame reconstruction and downsample the shallow features for the follow-up spatial-temporal aggregation.
They overlook that the input frames lose temporal correlations completely and thus too early temporal interactions could exaggerate artifacts.
To this end, we propose two modifications as shown in \cref{fig:baseline}: $i)$ use 2D RSTB to replace 3D CNN to disable temporal interactions, $ii)$ build skip connection between the shallow features and the upsampled refined features.
The ablation study is reported in \cref{tab:ablation1}, where \texttt{A+B(+Skip)} architecture represents that \texttt{A} and \texttt{B} are used for shallow feature extraction and feature-to-frame reconstruction respectively (with \texttt{Skip} connection). 
By disabling temporal interactions, (b) and (c) lead to a clear performance gain over widely-used (a), agreeing with the visualization in \cref{fig:feature}.
Except as shallow feature extractor, RSTB in (d) also show better performance in reconstructing high-fidelity frames than 3D CNN in (c).
With skip connection in (e), the fusion of shallow features and upsampled refined features could avoid the information loss caused by early spatial downsampling.

\subsubsection{Improvements in MSA.}
The proposed CSS-MSA is mainly powered by separable and cross-scale designs. we conduct the ablation study of them in \cref{tab:ablation2}. In the single-scale case, average pooling operator is discarded.
In the non-separable case, CSS-MSA is equivalent to cross-scale SW-MSA that performs 2D windowed MSA between normal query and average-pooled key and value.
As mentioned previously, separability introduces an inductive bias of paying more attention to spatial dimensions to harmonize with that informative clues concentrate on spatial dimensions instead of temporal dimension, namely the information skewness of video SCI.
Clearly, non-separability damages the performance while sacrificing the complexity reduction of factorized attentions~\cite{bertasius2021space,arnab2021vivit}.
Cross-scale interactions lead to a performance gain and a complexity reduction.
We further compare CSS-MSA with competitive MSAs for video SCI reconstruction in \cref{tab:ablation3}. As previously analyzed in \cref{tab:complexity}, G-MSA~\cite{dosovitskiy2020image} and F-MSA~\cite{arnab2021vivit,bertasius2021space} are impractical due to their quadratic computational complexity.
FW-MSA and SW-MSA are practical and they also are the sources of STFormer~\cite{wang2022spatial} and CTM-SCI~\cite{Zheng_2023_ICCV} respectively.
FW-MSA limits the spatial attention in TimeSformer~\cite{bertasius2021space} within local windows.
SW-MSA relaxes 3D window in W-MSA~\cite{liu2022video}
into spatial 2D window for temporal global receptive field.
Note that both FW-MSA and SW-MSA do not downsample key and value for cross-scale interactions.
Clearly, CSS-MSA outperforms them by a large margin for video SCI reconstruction.

\subsubsection{Improvements in FFN.}
The proposed GSM-FFN is powered by Gated Self-Modulation (GSM) and factorized Spatial-Temporal Convolution (STConv) that performs spatial aggregation and temporal aggregation in parallel and separately to enhance the locality. \cref{tab:ablation4} reports the ablation study on GSM-FFN. Without GSM, the channel expansion factor is set to $1$ to relieve the computational loads and parameters of STConv.
Clearly, GSM-FFN outperforms regular FFN by a large margin. 
After discarding GSM or replacing STConv with a 3D convolution, the resulting performance drops validate the superiority of GSM-FFN.

\section{Conclusion}
\label{sec:conclusion}
By analyzing the mixed degradation of spatial masking and temporal aliasing, we are the first to reveal the information skewness of video SCI, namely informative clues concentrate on spatial dimensions.
Previous works overlooks it and thus have the limited performance.
To this end, we tailor an efficient reconstruction architecture and Transformer block, dubbed HiSViT, to harmonize with the information skewness.
HiSViT captures long-range multi-scale spatial-temporal dependencies computationally efficiently.
Extensive experiments on grayscale, color, and real data demonstrate that our method achieves SOTA performance.

\section*{Acknowledgements}
This work was supported by the National Natural Science Foundation of China (grant number 62271414), Zhejiang Provincial Distinguished Young Scientist Foundation (grant number LR23F010001), Zhejiang ``Pioneer'' and ``Leading Goose'' R\&D Program (grant number 2024SDXHDX0006, 2024C03182), the Key
Project of Westlake Institute for Optoelectronics (grant number 2023GD007), the 2023 International Sci-tech Cooperation Projects under the purview of the ``Innovation Yongjiang 2035'' Key R\&D Program (grant number 2024Z126), Shanghai Municipal Science and Technology Major Project (2021SHZDZX0102), and the Fundamental Research Funds for the Central Universities.

%
%
\bibliographystyle{splncs04}
\bibliography{main}
\end{document}